\documentclass{article} % For LaTeX2e
\usepackage{iclr2027_conference,times}

\usepackage{amssymb}
\usepackage{amsmath}
\usepackage{amsthm}
\usepackage{amsfonts}
\usepackage{commath}
\usepackage{mathtools}

\usepackage{graphicx}
\usepackage{subcaption}
\usepackage[export]{adjustbox}
\usepackage{wrapfig}

\usepackage{multirow}

\usepackage{booktabs}

\usepackage[most]{tcolorbox}
\usepackage{algorithm}
\usepackage{algpseudocode}

\usepackage[colorinlistoftodos]{todonotes}

\usepackage{xcolor}
\usepackage{hyperref}

\definecolor{strat1}{HTML}{C0761A} % orange
\definecolor{strat2}{HTML}{3E9A48} % green
\definecolor{strat3}{HTML}{2F7FCC} % blue
\definecolor{strat4}{HTML}{8660C2} % purple
\definecolor{strat5}{HTML}{C2437E} % pink

\newcommand{\rainbowstratified}{\texorpdfstring{%
  \textcolor{strat1!85!white}{St}%
  \textcolor{strat2!85!white}{ra}%
  \textcolor{strat3!85!white}{ti}%
  \textcolor{strat4!85!white}{fi}%
  \textcolor{strat5!85!white}{ed}}{Stratified}}

\newcommand{\rev}[1]{#1}

\usepackage{setspace}

\usepackage[inline]{enumitem}

\newtheorem{definition}{Definition}
\newtheorem{prop}{Proposition}

\newtheorem{theorem}{Theorem}

\usepackage{amsmath,amsfonts,bm}

\def\eqref#1{equation~\ref{#1}}
\def\1{\bm{1}}

\DeclareMathAlphabet{\mathsfit}{\encodingdefault}{\sfdefault}{m}{sl}
\SetMathAlphabet{\mathsfit}{bold}{\encodingdefault}{\sfdefault}{bx}{n}

\usepackage{hyperref}
\usepackage{url}

\title{Learning \rainbowstratified{} Policies for \\Inference-Time Objectives}

\author{
Mridul Mahajan\textsuperscript{1},
Xuezhou Zhang\textsuperscript{1} \&
Aldo Pacchiano\textsuperscript{1,2}\\
\textsuperscript{1}Boston University\\
\textsuperscript{2}Broad Institute of MIT and Harvard\\
\texttt{\{mridulm,xuezhouz,pacchian\}@bu.edu}
}

\hypersetup{
    colorlinks=true,
    linkcolor=magenta!60!gray,
    citecolor=magenta!60!gray,
    urlcolor=magenta!60!gray
}

\iclrfinalcopy % Uncomment for camera-ready version, but NOT for submission.
\begin{document}

\maketitle
\lhead{Preprint}

% !TEX root =  main.tex
%%%%%%%%%%%%%%%%%%%%%%%%%%%%%%%%%%%%%
%%%%%%%%%%%%%%%%%%%%%%%%%%%%%%%%%%%%%
\begin{abstract}
Inference-time scaling improves the performance of large language models by sampling multiple responses and evaluating their collective utility. However, many existing approaches obtain these samples by repeatedly sampling from a single policy.
In this work, we study if the sample budget at inference time can instead be allocated across a population of policies.
We introduce stratified policies, where each of $k$ learned policies contributes one response to a $k$-sample inference-time objective.
We realize this policy population using separate LoRA adapters on a shared language model backbone and optimize it for pass@$k$.
We theoretically motivate the approach by analyzing the linear setting; a population of LoRAs can achieve zero best-of-$k$ approximation error in settings where every single LoRA has strictly positive prediction error.
Empirically, we demonstrate that stratified policies consistently improve pass@$k$ over a single-policy baseline across multiple reasoning tasks.
Overall, our results highlight how distributing samples across complementary policies can be a more effective use of sample budget than repeated i.i.d. sampling from a single policy at inference time.
\end{abstract}
% !TEX root = main.tex
%%%%%%%%%%%%%%%%%%%%%%%%%%%%%%%%%%%%%
%%%%%%%%%%%%%%%%%%%%%%%%%%%%%%%%%%%%%
\section{Introduction}
\label{sec:intro}

Inference-time computation is an effective way to improve the performance of large language models. Instead of relying on a single response, inference-time procedures sample multiple responses and evaluate their collective utility, such as best-of-$k$ \citep{bon}, reranking \citep{gsm8k}, or majority voting \citep{self_consistency}. Inspired by this, recent works explicitly optimize language models for the inference-time procedure used at deployment \citep{passk, bon_finetuning, pkpo}. This is different from standard post-training methods that optimize the quality of individual responses \citep{rlhf, dpo}.

These approaches share a common sampling scheme: all $k$ responses are sampled independently from a single policy. Although this is natural, it imposes a particular allocation of the inference-time budget: every response must come from the same distribution. Interestingly, even when this distribution assigns probability mass to several useful solution modes, i.i.d. sampling can repeatedly sample from the same modes and miss others. For example, consider a policy whose response distribution places all its probability mass on five distinct solution modes. From the AM-GM inequality, five i.i.d. samples can cover all five modes with probability at most $5!/5^5\approx3.84\%$. This upper bound is achieved when the policy assigns equal mass to each mode. Therefore, even a perfectly uniform policy may fail to realize this diversity in a finite set of responses.

\begin{figure*}[t]
    \centering
    \includegraphics[
        width=\textwidth
    ]{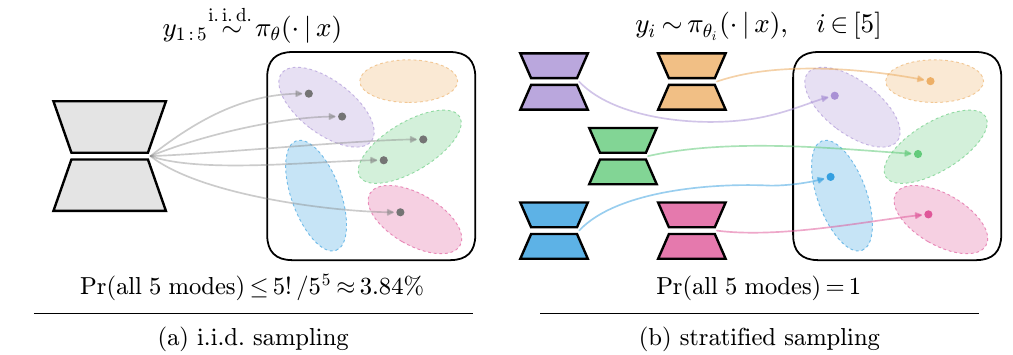}
    \caption{An illustrative example of stratification. With a fixed budget of five responses, i.i.d.\ sampling from a single policy may repeatedly sample the same solution modes (left). For five modes, the probability of covering all modes is at most $5!/5^5 \approx 3.84\%$, which is attained when the policy assigns equal probability to each mode. In contrast, stratified sampling allocates one response to each policy in a learned population, which allows specialized policies to produce complementary responses (right); under perfect specialization, all five modes are covered with probability one.}
    \label{fig:overview}
    \vspace{-0.5em}
\end{figure*}

In this work, we formulate inference-time optimization through a population of policies: the sample budget at inference time is treated as an allocation across a policy population. More concretely, given a budget of $k$ responses, we jointly learn $k$ policies and sample exactly one response from each. We refer to this as stratified sampling, inspired by classical stratified sampling, which partitions a population into strata and samples separately from each stratum \citep{stratified}. Unlike classical stratification, the learned policies need not partition the response space or have disjoint support. The standard single-policy formulation is recovered when all policies in the population are constrained to be identical. When this constraint is relaxed, the population can ``hedge its bets'' by learning policies that specialize in ways that improve the utility of the full set of responses. For pass@$k$, this can encourage policy complementarity without any explicit diversity objective (see Fig.~\ref{fig:overview}).

We realize this policy population efficiently using a shared (frozen) language model backbone with a separate low-rank adaptation (LoRA) \citep{lora} adapter for each policy. We focus on pass@$k$ as the utility function, where the objective is to maximize the probability that at least one response from the set is correct. Building on policy-gradient optimization for inference-time objectives, we train the population using Group Relative Policy Optimization (GRPO) \citep{grpo}.

Theoretically, we show that this population structure has an expressivity advantage in the linear setting. Specifically, when the target requires applying different linear maps over different regions of the input space, a LoRA population can assign a separate adapter to each region, but a single linear map must approximate all regions simultaneously. With sufficient rank, the policy population can achieve zero best-of-$k$ approximation error, whereas every single linear map incurs strictly positive error whenever the maps in different regions differ.

We evaluate stratified policies on a broad range of reasoning tasks spanning valid arithmetic expression construction, logical deduction, constraint satisfaction, and mathematical word problems. Across all tasks, stratification improves pass@$5$ over a single-policy baseline under the same inference-time response budget. On \textsc{Sudoku}, specifically, the pass@$5$ score increases from $74.2\%$ to $80.3\%$. Furthermore, when the rank of the single LoRA adapter is increased $5\times$ to match the population, performance remains essentially the same, at $74.2\%$, which suggests that the improvement is not explained by additional parameters alone. The learned population of policies also exhibits complementary behavior: for instance, on \textsc{Sudoku}, only $6.0\%$ of examples are solved by all five adapters, whereas $44.0\%$ are solved by exactly one. Together, these results demonstrate that population size provides an orthogonal scaling axis for inference-time objectives: instead of adding parameters to a single policy and repeatedly sampling from it, parameters can be distributed across policies that contribute complementary responses.

To summarize, our work makes the following contributions:
\begin{enumerate}[label={\Roman*.},leftmargin=*]
\item We introduce stratified policies for inference-time objectives, which replace $k$ i.i.d. samples from a single policy with one sample from each of $k$ learned policies (Section~\ref{sec:strat_pol}). For pass@$k$, the objective recovers the binary-reward special case of \citet{poppy}.
\item We provide a practical realization of this population using multiple LoRA adapters on a shared language model backbone, and train the resulting population using a group-mean pass@$k$ advantage within GRPO (Sections~\ref{sec:multi_lora}~and~\ref{sec:passk_learning}).
\item Theoretically, we prove that in a linear setting, stratification is strictly more expressive: with sufficient rank, a LoRA population can achieve zero best-of-$k$ approximation error while any single linear map incurs strictly positive error (Section~\ref{sec:theory}).
\item We empirically demonstrate consistent pass@$5$ improvements across multiple reasoning tasks, along with the emergence of complementary specialization (Section~\ref{sec:experiments}).
\end{enumerate}

We emphasize that our goal is not to design a state-of-the-art method for pass@$k$. Instead, we show that allocating the sample budget across a learned population of policies, rather than repeatedly sampling a single policy, is a complementary axis to single-policy pass@$k$ training.
% !TEX root =  main.tex
%%%%%%%%%%%%%%%%%%%%%%%%%%%%%%%%%%%%%
%%%%%%%%%%%%%%%%%%%%%%%%%%%%%%%%%%%%%
\section{Related Work}
\label{sec:rel_work}

\textbf{Optimizing Inference-Time Objectives.}
Sampling multiple responses can improve the performance of a language model by increasing the probability of finding a correct response \citep{repeated_sampling}. Methods such as verifiers and self-consistency then use these multiple responses to produce the final prediction \citep{gsm8k,self_consistency}. This motivates training for the utility of a response set instead of a single response. In line with this, recent work directly optimizes for the inference-time procedure used at deployment. Prior work in program synthesis already considered maximizing the expected reward of the best candidate program among multiple samples \citep{prog_syn}. For language models, inference-aware fine-tuning has been studied for best-of-$N$ sampling \citep{bon_finetuning}. \citet{passk} derive policy-gradient estimators for inference-time objectives such as pass@$k$ and majority voting. \citet{pkpo} propose a training method that can utilize more training samples than the inference-time budget. \citet{passk_training} also develop training advantages that are specific to pass@$k$ for reasoning models, but their group-relative estimator requires a training-time sample budget $N_{\text{rollout}}$ larger than the eventual pass@$k$ target $k$. This is different from the setting studied in our paper, where $N_{\text{rollout}}=k$, so we do not include this method as a baseline. \citet{advantage_shaping} interpret pass@$k$ advantage shaping as optimization of surrogate reward objectives. \citet{maxpo} study baselines for max@$k$ policy gradients and introduce a leave-two-out baseline for variance reduction. \citet{vpo} optimize sets of responses using randomized scalarizations of vector-valued rewards. These approaches optimize a single policy from which a set of responses is sampled. In contrast, our work jointly optimizes multiple policies and allocates one response to each.

\textbf{Learning Complementary Populations.}
Stochastic multiple choice learning trains deep ensembles through `winner-take-gradient' updates \citep{smcl}. \citet{lora_mcl} extend this to language modeling by training multiple LoRA adapters with a supervised winner-takes-all loss. In reinforcement learning, \citet{poppy} optimize the expected best reward from a population of policies, each parameterized by a shared encoder and separate decoders. Building on this, \citet{polynet} use a single (shared) decoder and condition it on strategy vectors. Our formulation considers stratification of general $k$-sample inference-time objectives; its pass@$k$ instantiation recovers the binary-reward special case of the population objective in \citet{poppy}.

\textbf{Structured Diversity in Language Models.}
\citet{upskill} learn distinguishable skills through a mutual-information objective and sample one response per skill at inference time. \citet{dqo} jointly optimize quality and semantic diversity using determinantal point processes over response embeddings. \citet{gcpo} reward responses based on their contribution to the group's semantic coverage, while \citet{darling} jointly reward response quality and semantic diversity. Several other methods explicitly encourage diversity through regularization, uniqueness rewards, diversity-aware preference optimization, or mode-conditioned reinforcement learning \citep{diversityaware,rare,dsdr,divpo,moda}. These methods explicitly encourage diversity; we instead jointly train separate LoRA policies using the inference-time objective, without an explicit diversity signal.

\textbf{Diversity at Inference Time.}
Diversity can also be encouraged during inference through diverse beam search, sampling without replacement, or prompt variation \citep{diversebeam,stochasticbeams,samplingdiversity}. \citet{robon} route test-time queries across existing language models. \citet{moa} aggregate responses through successive layers of language models. These methods modify inference-time generation, allocation, or aggregation, whereas we jointly train policies for their collective utility, with one response sampled from each policy.
% !TEX root =  main.tex
%%%%%%%%%%%%%%%%%%%%%%%%%%%%%%%%%%%%%
%%%%%%%%%%%%%%%%%%%%%%%%%%%%%%%%%%%%%
\section{Problem Setup}
\label{sec:setup}

\textbf{Language Models.}
We view a language model as a policy $\pi_\theta$, parameterized by $\theta$, that generates a response $y$ conditioned on a prompt $x$. Let $\rho$ denote the prompt distribution and $r(x,y)$ the reward function used to evaluate the response. A standard post-training objective maximizes the expected reward of a single response:
\[
\max_{\theta}\;
\mathbb{E}_{x\sim\rho,\;y\sim\pi_\theta(\cdot\mid x)}
\left[r(x,y)\right].
\]
\textbf{Inference-Time Objectives.}
At inference time, performance may be determined by the collective utility of a set of responses, instead of the quality of individual responses. Let $F(x,y_1,\ldots,y_k)$ denote the utility of $k$ responses under a given inference-time procedure. This is captured by the following training objective:
\[
\max_{\theta}\;
\mathbb{E}_{x\sim\rho,\;y_{1:k}\sim\pi_\theta(\cdot\mid x)^{\otimes k}}
\left[F(x,y_1,\ldots,y_k)\right],
\]
where the responses are sampled independently conditioned on $x$.

For instance, the utility function for best-of-$k$ is $F(x,y_1,\ldots,y_k) = \max_{i\in\{1,\ldots,k\}} r(x,y_i).$ When the reward is binary ($0$ for failure and $1$ for success), the expected best-of-$k$ utility exactly equals pass@$k$, which is the probability that at least one of the $k$ responses is correct.

\textbf{Objective.}
Given an inference-time utility $F$ and a sample budget $k$, our goal is to post-train a language model to maximize the expected utility of the set of responses. While our formulation captures general $k$-response utilities, we focus on pass@$k$ for the rest of the paper.

% !TEX root = main.tex
%%%%%%%%%%%%%%%%%%%%%%%%%%%%%%%%%%%%%
%%%%%%%%%%%%%%%%%%%%%%%%%%%%%%%%%%%%%
\section{Learning Stratified Policies}
\label{sec:framework}

In this section, we present a framework for learning a population of policies for inference-time objectives. First, we formulate inference-time optimization through stratified sampling from a population of policies (Section~\ref{sec:strat_pol}). We then describe a practical parameterization of this population using multiple LoRA adapters on a shared language model (Section~\ref{sec:multi_lora}). Finally, we instantiate our framework for pass@$k$ optimization (Section~\ref{sec:passk_learning}), and theoretically motivate it (Section~\ref{sec:theory}).

\subsection{Inference-Time Objectives through Stratified Sampling}
\label{sec:strat_pol}

As discussed in Section~\ref{sec:setup}, the single-policy formulation allocates a budget of $k$ responses by repeatedly sampling from a single policy $\pi_\theta$. However, repeatedly sampling solutions from the same policy may produce similar solutions, which could potentially limit their collective utility.

With this lens, we introduce a population of policies
$\Pi_\Theta=(\pi_{\theta_1},\ldots,\pi_{\theta_k})$, where
$\Theta=(\theta_1,\ldots,\theta_k)$. Instead of obtaining all $k$ responses from the same policy, we sample exactly one response from each policy. This gives us the following objective:
\begin{equation}
\max_{\Theta}\;
\mathbb{E}_{x\sim\rho,\;
y_{1:k}\sim\bigotimes_{i=1}^{k}\pi_{\theta_i}(\cdot\mid x)}
\left[F(x,y_1,\ldots,y_k)\right],
\label{eq:stratified_obj}
\end{equation}
where responses are sampled independently conditioned on $x$.

We refer to this (budget-)allocation of one sample per policy as stratified sampling. Note that the policies need not have disjoint support. Instead, each policy defines a response distribution that contributes one sample to the set. The single-policy objective can be recovered as the special case when
$\theta_1=\cdots=\theta_k$. Therefore, our formulation generalizes the single-policy setting.

Intuitively, optimizing a policy population allows each policy to contribute according to the utility of the full response set. When the objective rewards complementary responses (e.g., best-of-$k$), each policy can specialize in different subsets of prompts or different ways to solve them. Such specialization can emerge through the objective and learning dynamics, without a predefined partitioning of the prompt or response space. Although responses are sampled independently, the policies are trained jointly through the utility of the full response set. Therefore, each policy's learning signal depends on the responses generated by the other population members.

\subsection{A Practical Implementation with Multiple LoRA Adapters}
\label{sec:multi_lora}

With the stratified objective defined, we now describe a practical parameterization of the policy population. A straightforward approach is to maintain $k$ independent language models, one for each policy. However, this requires storing and training a separate set of parameters for each policy in the population. Instead, we utilize a frozen language model backbone and parameterize each policy using a separate low-rank adaptation (LoRA) module \citep{lora}.

Let $W_\ell$ denote a frozen weight matrix, where $\ell$ indexes the matrices to which LoRA is applied. For policy $i$, the corresponding adapted matrix is:
\begin{equation}
W_\ell+\Delta W_\ell^{(i)},
\qquad
\operatorname{rank}\!\left(\Delta W_\ell^{(i)}\right)\le R,
\label{eq:multi_lora}
\end{equation}
where $R$ denotes the LoRA rank. In practice, the update is parameterized as:
\begin{equation}
\Delta W_\ell^{(i)} = \frac{\alpha}{R}\, B_\ell^{(i)} A_\ell^{(i)},
\label{eq:lora_factors}
\end{equation}
where $A_\ell^{(i)}$ and $B_\ell^{(i)}$ are the LoRA factors and $\alpha$ is a scaling parameter (set to $R$ in our experiments).

All policies share the same frozen matrices $W_\ell$ but use separate LoRA adapters. Given a prompt, we generate one response per adapter and evaluate the resulting set according to utility $F$. This parameterization realizes a population without maintaining $k$ full model copies.

\subsection{Instantiation for Pass@$k$}
\label{sec:passk_learning}

We now instantiate our framework for the best-of-$k$ utility: $F(x,y_1,\ldots,y_k)=\max_i r(x,y_i)$.
For binary correctness rewards, the objective in Eq.~\ref{eq:stratified_obj} is equal to the probability that at least one policy in the population generates a correct response.

\textbf{Learning with Leave-One-Out Baseline.}
We build on the inference-time policy gradient formulation of \citet{passk}.
When using a leave-one-out (LOO), response $y_i$ is assigned an advantage equal to its marginal contribution to the maximum reward:
\begin{equation}
A_i^{\mathrm{LOO}}
=
\max_{j=1,\ldots,k}r_j
-
\max_{j\ne i}r_j,
\label{eq:loo_advantage}
\end{equation}
where $r_i=r(x,y_i)$ and $k\ge 2$. The advantage is non-zero only for the best response, and is effectively equal to the difference between the highest and second-highest rewards.

Policy gradients for our population of policies can be constructed similarly since responses are sampled independently conditioned on the prompt. More concretely, for policy $\pi_{\theta_i}$, the (unclipped) policy gradient is given by:
\begin{equation}
\nabla_{\theta_i}J_{\max}(\Theta)
=
\mathbb{E}
\left[
A_i^{\mathrm{LOO}}\,
\nabla_{\theta_i}\log\pi_{\theta_i}(y_i\mid x)
\right],
\label{eq:stratified_loo_gradient}
\end{equation}
where $J_{\max}$ denotes Eq.~\ref{eq:stratified_obj} with the maximum-reward utility. The baseline $\max_{j\ne i}r_j$ does not depend on $y_i$. Consequently, subtracting it does not change the policy gradient. In practice, we use this advantage construction within Group Relative Policy Optimization (GRPO) \citep{grpo}.

\textbf{Denser Learning Signal via Group-Mean Baseline.}
When rewards are binary, the LOO advantage is zero whenever multiple responses in the group are successful, even if other responses fail. Thus, these groups provide no learning signal even if there is room for improvement. Motivated by this, we instead use a group-mean baseline for a denser learning signal. More concretely, we compute group mean $\bar r=\frac{1}{k}\sum_{j=1}^{k}r_j$, and sample the winner $i^\star$ uniformly at random from $\arg\max_{i\in\{1,\ldots,k\}} r_i$ to randomly break ties. Response $y_i$ is then assigned advantage:
\begin{equation}
A_i
=
\begin{cases}
r_i-\bar r, & \text{if } i=i^\star,\\
0, & \text{otherwise}.
\end{cases}
\label{eq:mean_wta_advantage}
\end{equation}

For binary rewards, if $m$ of the $k$ responses are correct, the selected winner receives advantage $1-m/k$. Thus, every group containing both correct and incorrect responses produces a non-zero update, whereas groups that are all-correct or all-incorrect produce no update. Randomly breaking ties prevents favoring a particular policy in the population. Appendix~\ref{sec:app.baseline} shows that, on \textsc{Sudoku}, group-mean improves population performance, and stratification helps under both estimators. Unlike LOO, this modification does not give an unbiased estimate of the best-of-$k$ policy gradient. \rev{Appendix~\ref{sec:app.maxrl} discusses the objective that it optimizes and shows its connection to \textsc{MaxRL} \citep{maxrl}.}

\subsection{Theoretical Analysis}
\label{sec:theory}

\begin{wrapfigure}{r}{0.47\textwidth}
  \centering
  \vspace{-1.7em}
  \includegraphics[width=\linewidth]{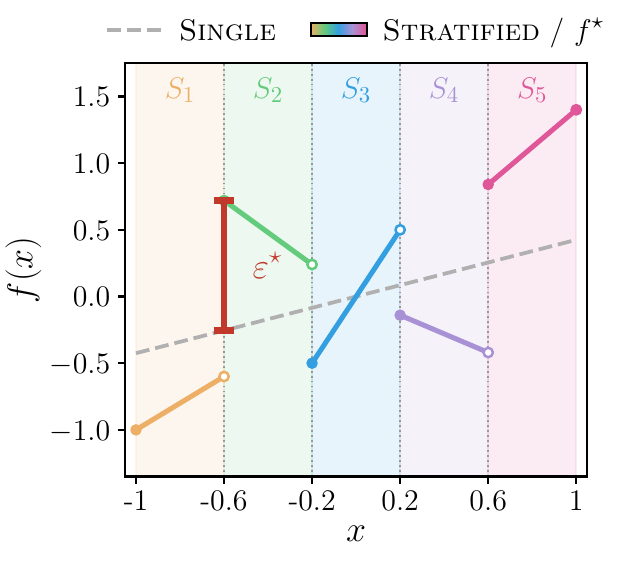}
  \vspace{-1.8em}
  \caption{An illustration of Theorem~\ref{thm:expressivity}. Colored segments represent the regional target maps $f^\star(x)=T_jx$. A rank-1 population fits every region exactly, whereas the best single linear map $Bx$ has an irreducible gap $\varepsilon^\star>0$.}
  \label{fig:theory}
  \vspace{-1em}
\end{wrapfigure}

We theoretically demonstrate that a stratified population of policies can be more expressive than a single policy for best-of-$K$ evaluation. More concretely, we consider a linear setting with a frozen base model $Wx$, where $W\in\mathbb{R}^{d\times d}$. Inputs lie in the unit ball $\mathbb{B}_2^d \coloneqq \left\{x\in\mathbb{R}^d : \|x\|_2 \le 1\right\}$, which is partitioned into regions $S_1,\ldots,S_K$. The target function $f^\star$ is piecewise linear on the partition:
\[
f^\star(x)=T_jx, \qquad \forall x\in S_j,
\]
where $T_j\in\mathbb{R}^{d\times d}$ is the target map for region $S_j$.

Policy $i$ in the population updates the base model with a rank-$r$ LoRA adapter $\Delta_i$, which gives us the predictor $(W+\Delta_i)x$. Evaluation is best-of-$K$: the population error at an input $x$ is determined by the best prediction among its member policies.

Intuitively, a rank-$r$ LoRA population can assign each adapter to a regional correction $T_j-W$. With sufficient rank, this achieves zero error. In contrast, a single linear predictor has to approximate all maps simultaneously. Fig.~\ref{fig:theory} illustrates this gap in a one-dimensional example with $K=5$.

\begin{definition}[Prediction errors]
For a fixed population of adapters $\Delta_{1:K}$, we define the best-of-$K$ prediction error at input $x$ as
\[
\varepsilon_{\mathrm{pop}}(x;\Delta_{1:K})
\coloneqq
\min_{i\in[K]}
\|(W+\Delta_i)x-f^\star(x)\|_2.
\]
For a single linear predictor $B$, we define
\[
\varepsilon_{\mathrm{single}}(x;B)
\coloneqq
\|Bx-f^\star(x)\|_2.
\]
\end{definition}

While the population is fixed across inputs, the minimizing policy in $\varepsilon_{\mathrm{pop}}$ may vary with $x$. Thus, to make accurate best-of-$K$ predictions, at least one policy in the population must predict well at each input. This does not require any individual policy to approximate the target over the entire domain $\mathbb{B}_2^d$. Note that the minimum represents best-of-$K$ evaluation; we are not learning a routing rule.

We use $\sigma_{\ell}(A)$ to denote the $\ell$-th largest singular value of a matrix $A$. When $\ell>\operatorname{rank}(A)$, $\sigma_{\ell}(A)$ is defined as zero.

We build on the expressivity result in \citet{lora_expressivity} and formalize this separation below; the proof is provided in Appendix~\ref{sec:app.proofs}.

\begin{theorem}[Expressive power of LoRA populations for best-of-$K$]
Suppose each region $S_j$ contains a closed Euclidean ball of radius $\rho>0$. Then, we have the following minimax guarantee:
\[
\min_{\substack{\Delta_{1:K}\\
\operatorname{rank}(\Delta_i)\le r,\ \forall i\in[K]}}
\max_{x\in\mathbb{B}_2^d}
\varepsilon_{\mathrm{pop}}(x;\Delta_{1:K})
\le
\max_{j\in[K]}
\sigma_{r+1}(T_j-W),
\]
whereas
\[
\min_{B\in\mathbb{R}^{d\times d}}
\max_{x\in\mathbb{B}_2^d}
\varepsilon_{\mathrm{single}}(x;B)
\ge
\frac{\rho}{2}
\max_{i,j\in[K]}
\|T_i-T_j\|_{\mathrm{op}}.
\]
Consequently, if $r\ge\max_j\operatorname{rank}(T_j-W)$ and the maps $\{T_j\}_{j=1}^K$ are not all identical, then a LoRA population can achieve zero error, while a single linear map always has strictly positive error.
\label{thm:expressivity}
\end{theorem}

It is important to note that we cannot eliminate this separation by simply increasing the LoRA rank, since the lower bound on single-predictor error holds regardless of rank.

While this analysis does not provide any guarantees for nonlinear language models or for learned population of policies, it motivates our use of multiple adapters for pass@$k$.
% !TEX root =  main.tex
%%%%%%%%%%%%%%%%%%%%%%%%%%%%%%%%%%%%%
%%%%%%%%%%%%%%%%%%%%%%%%%%%%%%%%%%%%%
\newpage

\section{Experimental Evaluation}
\label{sec:experiments}

\begin{figure*}[t]
    \centering
    \includegraphics[
        width=\textwidth
    ]{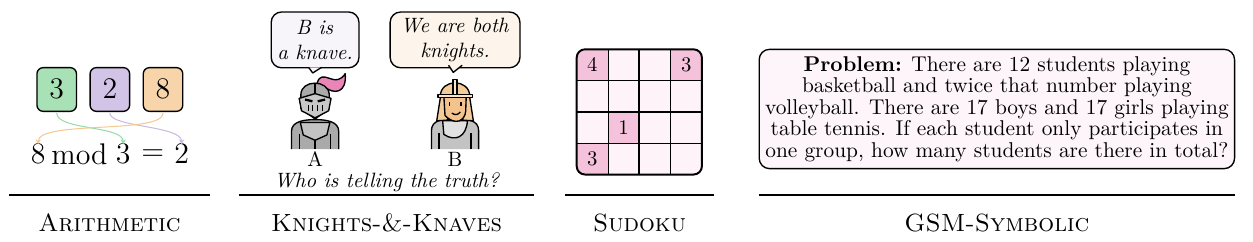}
    \caption{We evaluate our method on a diverse set of reasoning tasks. These span constructing valid arithmetic expressions, logical deduction, constraint satisfaction, and mathematical word problems.}
    \label{fig:reasoning_tasks}
\end{figure*}
% !TEX root =  main.tex
\begin{figure*}[t]
    \centering
    \includegraphics[
        width=\textwidth
    ]{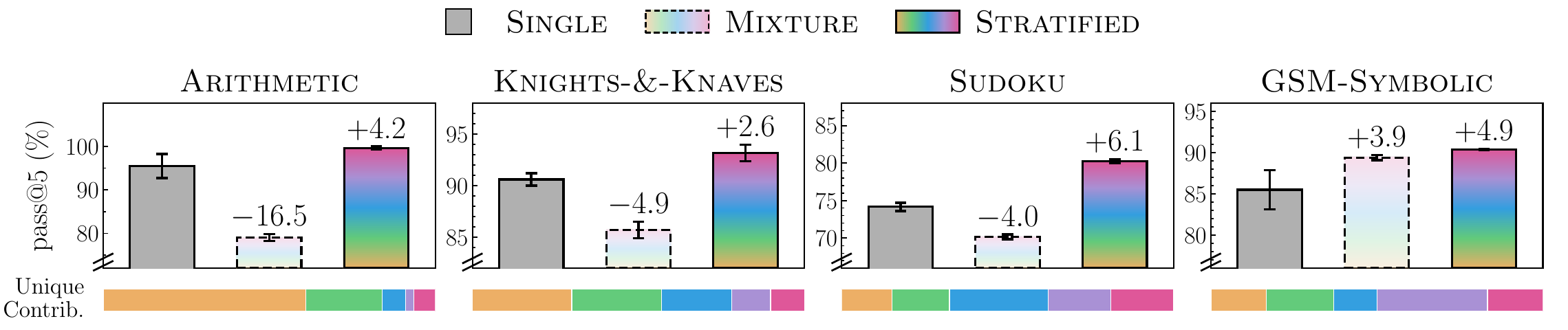}
    \caption{Pass@5 accuracy of a single policy (\textsc{Single}) versus our stratified population of five specialized policies (\textsc{Stratified}) across reasoning tasks. \textsc{Mixture} (dashed) samples from the same five policies i.i.d.\ instead of once from each. Error bars denote standard error across three seeds. Each panel's $y$-axis is truncated (kink mark) to make the gap visible. The (renormalized) horizontal bar below each panel shows how much each policy contributes to the population’s unique-only solutions, i.e., problems solved by exactly one policy. \textsc{Stratified} outperforms \textsc{Single} on pass@5 across all four tasks, and each policy member contributes unique solutions to the population.}
    \label{fig:passk_scores}
    \vspace{-0.5em}
\end{figure*}

In this section, we empirically evaluate stratified policies to address the following research questions:
\begin{enumerate*}[label=(\roman*)]
    \item Does sampling from a stratified policy population at inference time improve pass@$k$ over repeatedly sampling from a single policy?
    \item Do the learned policies develop complementary behavior, i.e., solve different subsets of problems?
    \item Can the gains from policy stratification be explained simply by additional trainable parameters, or does it depend on maintaining a policy population?
\end{enumerate*}

We begin by describing the benchmark tasks and experimental setup. Next, we present the main performance results, followed by analyses of policy complementarity and ablations.

%%%%%%%%%%%%%%%%%%%%%%%%%%%%%%%%%%%%%%%%%%%%%%%
\subsection{Experimental Setup}
\label{sec:exp_setup}

\textbf{Tasks.}
We evaluate on four reasoning tasks (Fig.~\ref{fig:reasoning_tasks}): \textsc{Arithmetic}, introduced in \citet{upskill}, in which the model is given three unordered digits and must form a valid expression that relates them; \textsc{Knights-\&-Knaves}, a challenging logical deduction task used to evaluate LLM reasoning and generalization \citep{knights_knaves}; \textsc{Sudoku}, a constraint-satisfaction task on which LLMs continue to struggle \citep{sudoku_bench}; and \textsc{GSM-Symbolic}, a procedurally generated variant of GSM8K designed to test mathematical reasoning more reliably, on which many models perform worse than on the original GSM8K problems \citep{gsm_symbolic}. We use the \textsc{Reasoning Gym} implementations \citep{reasoning_gym}, which is a collection of reasoning environments built to standardize the training and evaluation of reasoning models. See Appendix~\ref{sec:app.task_details} for more details.

\textbf{Training.}
For \textsc{Arithmetic}, we train a \texttt{TinyGPT} causal transformer. For \textsc{Knights-\&-Knaves} and \textsc{Sudoku}, we fine-tune Qwen2.5-1.5B-Instruct \citep{qwen}, while for \textsc{GSM-Symbolic}, we fine-tune Llama-3.1-8B-Instruct \citep{llama}. We use rank-$32$ LoRA adapters and train with GRPO without a KL penalty, using the advantage defined in Eq.~\ref{eq:mean_wta_advantage}. The stratified population (\textsc{Stratified}) maintains $5$ policies and samples one response from each, whereas the single-policy baseline (\textsc{Single}) trains one policy using the same five-sample budget and advantage, which is similar to \citet{passk} without stratification. Both of these methods assign nonzero advantage to at most one response. So only one needs a gradient-tracked forward and backward pass. Therefore, both require the same order of training compute. See Appendix~\ref{sec:app.training_details} for more details.

\textbf{Evaluation.}
We report pass@$5$ under an inference-time budget of five samples. Results are averaged over three random seeds, and error bars denote standard error. To check whether sampling exactly one response from each adapter is essential, we also compare against \textsc{Mixture}. It uses the same trained adapters, but samples an adapter uniformly at random for each of the five responses, which misses several adapters with high probability (see Fig.~\ref{fig:overview}). See Appendix~\ref{sec:app.eval_details} for more details.

% !TEX root =  main.tex
\begin{figure*}[t]
    \centering
    \includegraphics[
        width=\textwidth
    ]{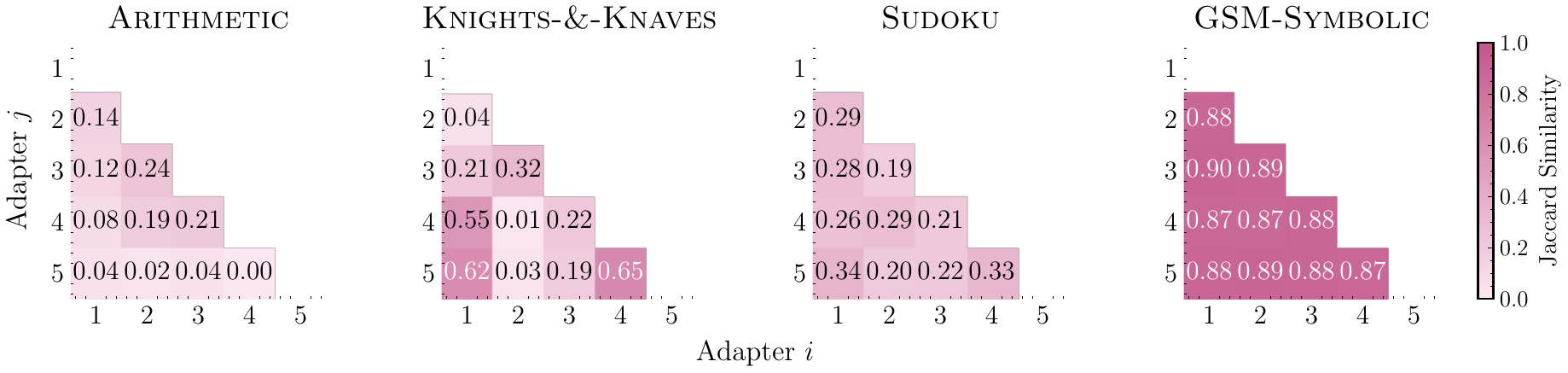}
    \vspace{-2em}
    \caption{Pairwise Jaccard similarity between the sets of problems solved by each policy. Lower values indicate that a pair of policies solve mostly disjoint subsets of problems; a value of $1$ indicates total overlap. \textsc{Arithmetic}, \textsc{Knights-\&-Knaves}, and \textsc{Sudoku} show low overlap; the values for \textsc{GSM-Symbolic} remain $\leq 0.9$, which still shows non-redundancy in this harder setting.}
    \label{fig:jaccard_heatmaps}
\end{figure*}

% !TEX root =  main.tex
\begin{figure*}[t]
    \centering
    \vspace{-0.5em}
    \includegraphics[
        width=\textwidth
    ]{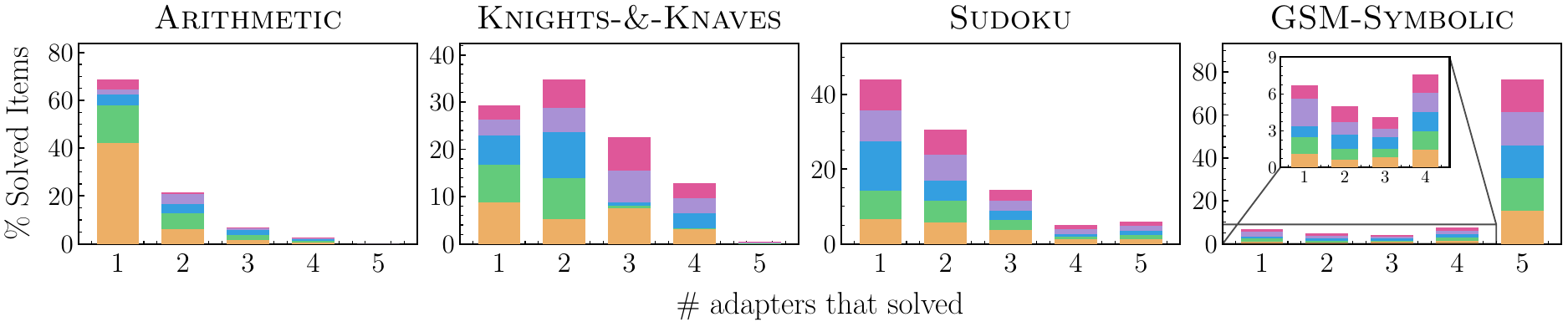}
    \vspace{-1.8em}
    \caption{\rev{Distribution, over solved problems, of the number of policies solving each problem under greedy decoding. Colors show the contribution of each policy. For \textsc{Arithmetic}, \textsc{Knights-\&-Knaves}, and \textsc{Sudoku}, mass is concentrated at low counts; this indicates that most solved problems are solved by only one or two policies (not the full population). For \textsc{GSM-Symbolic}, there is substantial mass on counts below $5$; this shows that not all problems are solved by all $5$ policies.}}
    \label{fig:unique_solves}
    \vspace{-1.3em}
\end{figure*}

\vspace{-0.8em}

%%%%%%%%%%%%%%%%%%%%%%%%%%%%%%%%%%%%%%%%%%%%%%%
\subsection{Results}
\label{sec:main_results}

\textbf{Stratification improves pass@$5$.}
Across all four tasks, \textsc{Stratified} improves pass@$5$ over \textsc{Single} under the same inference-time response budget (see Fig.~\ref{fig:passk_scores}). If we view it as reduction in the gap to optimality ($100\%$ pass@$5$), \textsc{Stratified} reduces the gap by $92\%$ on \textsc{Arithmetic}, and between $24$ and $34\%$ on the other three tasks. \textsc{Mixture} is worse than \textsc{Stratified} on every task and worse than \textsc{Single} on three out of four, which suggests that all adapters are needed to improve pass@$5$. On pass@$1$ (see Appendix~\ref{sec:app.full_results}), however, \textsc{Single} outperforms \textsc{Stratified} on most tasks. This trade-off between pass@$1$ and pass@$5$ is not surprising because we are optimizing for pass@$5$ directly \citep{why_passk}, not any individual policy's pass@$1$.

\textbf{Policies make unique contributions.}
The gains in pass@$5$ from stratification suggest that the learned policies are not redundant copies of one another. Fig.~\ref{fig:passk_scores} additionally shows the (normalized) share of each policy's unique-only contribution per task under greedy decoding. Across all four tasks, each policy in the population solves problems that cannot be solved by the rest of the population. Therefore, the entire population (and not a single policy) gives complementary coverage.

\textbf{Policies solve different problem sets.}
We analyze if different policies solve different subsets of problems. Fig.~\ref{fig:jaccard_heatmaps} shows pairwise Jaccard similarity (defined for sets $A$ and $B$ as $|A\cap B|/|A\cup B|$) between the set of items solved by the learned policies under greedy decoding. The overlap is lowest on \textsc{Arithmetic} and \textsc{Knights-\&-Knaves} (several pairs with similarity below $0.1$), fairly low on \textsc{Sudoku}, and higher (though still $\leq 0.9$) on \textsc{GSM-Symbolic}. Interestingly, \textsc{Knights-\&-Knaves} shows a bimodal pattern, where some pairs are almost identical and others complementary.

\textbf{Most solved problems are solved by only one or two policies.}
The distribution over how many of the five policies solve a given problem (see Fig.~\ref{fig:unique_solves}) is concentrated at low counts for \textsc{Arithmetic}, \textsc{Knights-\&-Knaves}, and \textsc{Sudoku}. This indicates that most solved problems are not solved by the entire population, but only by one or two policies. For \textsc{GSM-Symbolic} instead, although there is substantial mass at count $5$, a non-trivial fraction of problems are still solved by fewer than all five policies. This shows that part of the gain in pass@$5$ comes from complementary coverage.

% !TEX root =  main.tex
\begin{figure*}[t]
    \centering
    \includegraphics[width=\textwidth]{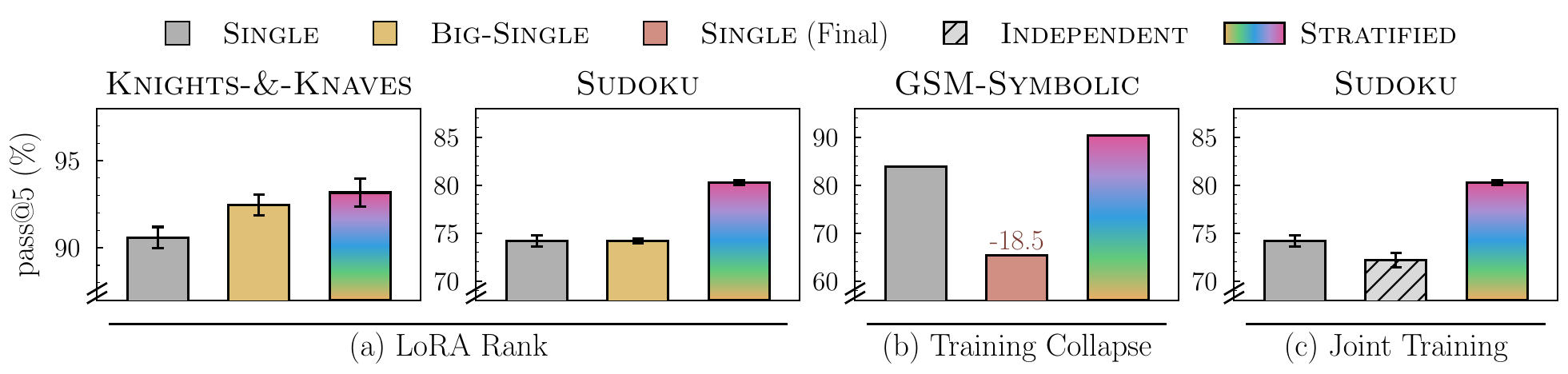}
    \caption{(a) Scaling the rank of \textsc{Single} by $5\times$ (\textsc{Big-Single}) to match \textsc{Stratified}'s total trainable parameters still underperforms it. (b) \textsc{Single} is unstable on \textsc{GSM-Symbolic}; the final checkpoint, \textsc{Single} (Final), achieves a substantially lower score than the pre-collapse checkpoint used elsewhere. \textsc{Stratified} does not exhibit this collapse on any seed. (c) An independently trained ensemble (\textsc{Independent}) with matched total training compute underperforms \textsc{Stratified}.}
    \label{fig:additional_results}
\end{figure*}
% !TEX root =  main.tex
\begin{figure*}[t]
    \centering
    \vspace{-1em}
    \includegraphics[width=\textwidth]{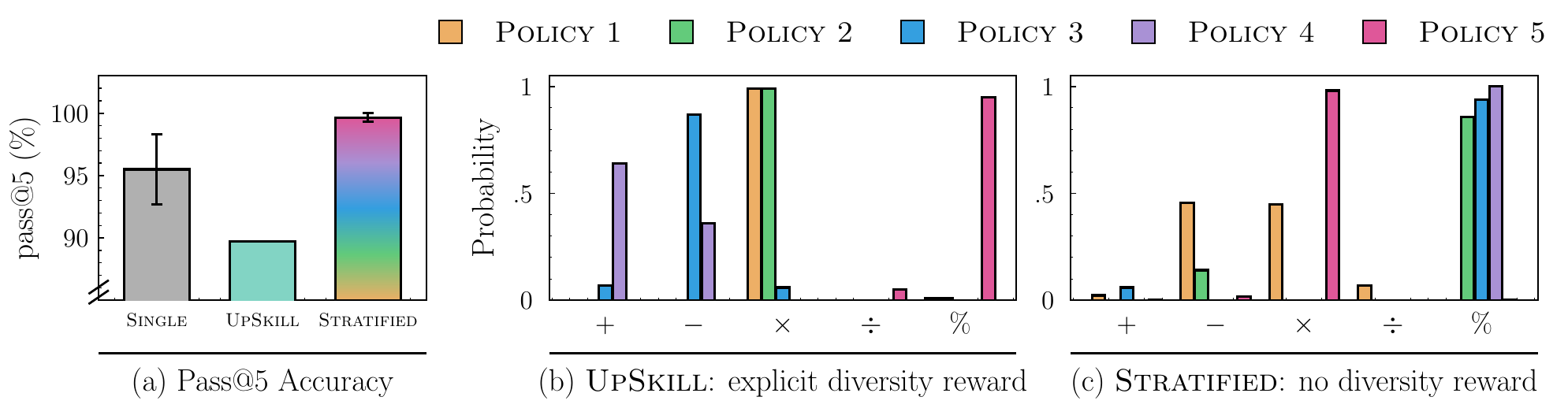}
    \caption{Comparison with \textsc{UpSkill} on \textsc{Arithmetic}. (a) \textsc{Stratified} outperforms \textsc{UpSkill}, which uses an explicit diversity reward. (b, c) Operator distribution of each policy. Stratified policies achieve a comparable level of diversity to \textsc{UpSkill}: different policies focus on different operators.}
    \label{fig:operator_preferences}
    \vspace{-1.5em}
\end{figure*}

\textbf{Stratified policies provide a new scaling axis.}
To test whether the gains from \textsc{Stratified} come from more trainable parameters, we compare it against \textsc{Big-Single}, which increases the LoRA rank of \textsc{Single} by $5\times$ to match the total number of trainable parameters in \textsc{Stratified}. Fig.~\ref{fig:additional_results}~(a) shows that \textsc{Big-Single} improves pass@$5$ from $90.6$ to $92.5$ on \textsc{Knights-\&-Knaves}, but achieves no gains on \textsc{Sudoku}. In both cases, \textsc{Stratified} still performs better. Therefore, the improvement from stratification cannot be explained by additional parameters alone. This shows that population size could be an orthogonal scaling axis to simply increasing the rank of a single adapter. An independently trained ensemble (\textsc{Independent}) of five \textsc{Single} runs with matched total training compute is also worse on \textsc{Sudoku} (Fig.~\ref{fig:additional_results}~(c)), which shows that the gains depend on joint training.

\textbf{Stratified policies can be robust to training collapse.}
We found \textsc{Single} to be unstable late in training on \textsc{GSM-Symbolic} on $2$ out of $3$ seeds. For instance, the final checkpoint for one of the seeds achieves only $65.4$ pass@$5$, while the pre-collapse checkpoint used elsewhere in the paper achieves $83.9$ (Fig.~\ref{fig:additional_results}~(b) shows one such seed). This corresponds to a drop of $18.5$ points. In contrast, \textsc{Stratified} was stable across all three seeds. This shows that stratified learning can be more stable than optimizing a single policy for pass@$5$.

\textbf{Stratified policies specialize without an explicit diversity reward.}
\textsc{UpSkill} \citep{upskill} encourages diverse responses using an explicit mutual-information diversity reward. \textsc{Stratified} instead learns to improve collective population performance using the pass@$5$ advantage defined in Eq.~\ref{eq:mean_wta_advantage}, with diversity useful insofar as it improves this objective. We additionally compare against \textsc{UpSkill} on \textsc{Arithmetic}, the benchmark introduced in their paper. Interestingly, as can be seen in Fig.~\ref{fig:operator_preferences}~(a), even \textsc{Single} achieves higher pass@$5$ than \textsc{UpSkill} on this task, despite using no explicit diversity reward. Importantly, not optimizing for diversity does not prevent it from emerging. Fig.~\ref{fig:operator_preferences}~(b, c) shows that the policies learned by \textsc{Stratified} develop different operator preferences despite having no explicit diversity reward. Although three of the five policies primarily use the modulo operator, they are still complementary: among correct responses using the \% operator, their pairwise Jaccard similarity ranges from $0.12$ to $0.23$, and $32.8\%$ of the full set of evaluation problems are solved using \% by exactly one of these three policies.
% !TEX root =  main.tex
%%%%%%%%%%%%%%%%%%%%%%%%%%%%%%%%%%%%%
%%%%%%%%%%%%%%%%%%%%%%%%%%%%%%%%%%%%%
\section{Concluding Discussion}
\label{sec:conclusions}

In this work, we introduced a stratified formulation of inference-time objectives, where the response budget is allocated across a population of jointly learned policies. Using separate low-rank adapters on a shared language model backbone, we empirically demonstrated improvements in pass@$k$ over a single-policy baseline across four reasoning tasks under the same inference-time budget of responses. While our formulation captures general $k$-sample objectives, our experiments focus on pass@$k$. A promising direction for future work is to study other inference-time objectives, such as majority voting and verifier-based selection. Further, adapting the population size and sample allocation to the prompt and available inference budget is another interesting direction.

%%%%%%%%%%%%%%%%%%%%%%%%%%%%%%%%%%%%%%%%%%%%%%%%%%%%%%%%%%%%%%%%
%% NOTE: THIS MARKS THE END OF THE "MAIN TEXT"
%%%%%%%%%%%%%%%%%%%%%%%%%%%%%%%%%%%%%%%%%%%%%%%%%%%%%%%%%%%%%%%%

%%%%%%%%%%%%%%%%%%%%%%%%%%%%%%%%%%%%%%%%%%%%%%%%%%%%%%%%%%%%%%%%
%% Bibliography
%%%%%%%%%%%%%%%%%%%%%%%%%%%%%%%%%%%%%%%%%%%%%%%%%%%%%%%%%%%%%%%%
\bibliography{main}
\bibliographystyle{iclr2027_conference}

\newpage

% !TEX root =  main.tex
%%%%%%%%%%%%%%%%%%%%%%%%%%%%%%%%%%%%%
%%%%%%%%%%%%%%%%%%%%%%%%%%%%%%%%%%%%%
\appendix
{
    \allowdisplaybreaks
    % !TEX root =  main.tex
%%%%%%%%%%%%%%%%%%%%%%%%%%%%%%%%%%%%%
%%%%%%%%%%%%%%%%%%%%%%%%%%%%%%%%%%%%%
\section{Table of Contents}
\label{sec:app.toc}

In this section, we provide an outline of the contents provided in the paper's appendices.
\begin{itemize}[leftmargin=*]
    \item Appendix~\ref{sec:app.proofs} contains the proof for Theorem~\ref{thm:expressivity}.
    \item Appendix~\ref{sec:app.task_details} describes the tasks used in the experiments.
    \item Appendix~\ref{sec:app.training_details} provides additional training details for the experiments.
    \item Appendix~\ref{sec:app.eval_details} provides additional evaluation details for the experiments.
    \item Appendix~\ref{sec:app.full_results} provides the full results for the main experiments, including pass@$1$ scores.
    \rev{\item Appendix~\ref{sec:app.baseline} empirically justifies the choice of the group-mean baseline over leave-one-out.}
    \rev{\item Appendix~\ref{sec:app.complementarity} provides further analysis of the learned complementarity.}
    \rev{\item Appendix~\ref{sec:app.population_size} shows that stratified policies improve performance across different population sizes.}
    \rev{\item Appendix~\ref{sec:app.maxrl} shows the connection between our method and maximum likelihood RL.}
\end{itemize}

    % !TEX root =  main.tex
%%%%%%%%%%%%%%%%%%%%%%%%%%%%%%%%%%%%%
%%%%%%%%%%%%%%%%%%%%%%%%%%%%%%%%%%%%%
\section{Proofs}\label{sec:app.proofs}
In this section, we provide the proof of Theorem~\ref{thm:expressivity}.

For each region $S_j$, the population can adapt the base model parameters $W$
to the target map $T_j$. Lemma~1 with $L=1$ in \citet{lora_expressivity} tells us that there exists a rank-$r$ adapter $\Delta_j$ such that:
\[
\|W+\Delta_j-T_j\|_{\mathrm{op}}
=
\sigma_{r+1}(T_j-W).
\]
Therefore, for all $x\in S_j$, the following holds:
\begin{align*}
\min_{i\in[K]}
\|(W+\Delta_i)x-f^\star(x)\|_2
&\le
\|(W+\Delta_j)x-T_jx\|_2 \\
&\le
\|W+\Delta_j-T_j\|_{\mathrm{op}}\|x\|_2 \\
&\le
\sigma_{r+1}(T_j-W).
\end{align*}
The last inequality follows from $\|x\|_2\le1$.

Since the unit ball $\mathbb{B}_2^d$ is partitioned into $S_1,\ldots,S_K$, taking the
maximum over $x$ gives us:
\[
\min_{\operatorname{rank}(\Delta_i)\le r}
\max_{\|x\|_2\le1}\min_{i\in[K]}
\|(W+\Delta_i)x-f^\star(x)\|_2
\le
\max_{j\in[K]}
\sigma_{r+1}(T_j-W).
\]

We now derive a lower bound for a single linear predictor. Fix matrix $B$ and define:
\[
\varepsilon
\coloneqq
\max_{\|x\|_2\le1}
\|Bx-f^\star(x)\|_2.
\]
Let $c_j$ denote the center of a closed Euclidean ball of radius $\rho$ contained in $S_j$. Then, for every unit vector $u$, $c_j+\rho u,c_j-\rho u\in S_j$.
Since $f^\star(x)=T_jx$ on $S_j$,
\[
\|(B-T_j)(c_j+\rho u)\|_2\le\varepsilon,
\qquad
\|(B-T_j)(c_j-\rho u)\|_2\le\varepsilon.
\]
Hence, by the triangle inequality, we have:
\begin{align*}
\|(B-T_j)u\|_2
&=
\frac{1}{2\rho}
\left\|
(B-T_j)(c_j+\rho u)
-
(B-T_j)(c_j-\rho u)
\right\|_2 \\
&\le
\frac{\varepsilon}{\rho}.
\end{align*}
Taking the maximum over all unit vectors $u$ gives us:
\[
\|B-T_j\|_{\mathrm{op}}
\le
\frac{\varepsilon}{\rho}.
\]
Therefore, for any $i,j \in [K]$,
\[
\|T_i-T_j\|_{\mathrm{op}}
\le
\|T_i-B\|_{\mathrm{op}}
+
\|B-T_j\|_{\mathrm{op}}
\le
\frac{2\varepsilon}{\rho}.
\]
Consequently, the following holds:
\[
\varepsilon
\ge
\frac{\rho}{2}
\max_{i,j\in[K]}
\|T_i-T_j\|_{\mathrm{op}}.
\]
Taking the minimum over $B$ gives us:
\[
\min_B
\max_{\|x\|_2\le1}
\|Bx-f^\star(x)\|_2
\ge
\frac{\rho}{2}
\max_{i,j\in[K]}
\|T_i-T_j\|_{\mathrm{op}}.
\]

This completes the proof.
    % !TEX root =  main.tex
%%%%%%%%%%%%%%%%%%%%%%%%%%%%%%%%%%%%%
\section{Task Details}\label{sec:app.task_details}

\subsection{\textsc{Arithmetic}}
This environment was introduced by \citet{upskill} and requires the model to combine three given single-digit numbers into a valid arithmetic expression that relates them.

An instance of the task is an unordered triple of digits $(n_1, n_2, n_3) \in \{0, \dots,9\}^3$, which is constructed such that some pair of the three, when combined via one of the operators $\{+, -, \times, \div, \%\}$, gives the third. For example, the triple $\{2, 3, 6\}$ can be used to form the expression $2 \times 3 = 6$. The three digits are presented to the model in random order. Therefore, a valid order is not revealed by their position. Task variability comes from the sampled digits, the operator used to construct the triple, and the order in which the digits are shown.

The model needs to respond with an expression of the form $L \; \mathrm{op} \; R$, where $L$ and $R$ are two of the three given digits and $\mathrm{op}$ is one of the operators. A task is considered solved iff $L \; \mathrm{op} \; R$ gives the remaining digit.

\subsection{\textsc{Knights-\&-Knaves}}
This is a logical-deduction task in which a model is given a set of self-referential statements and must determine which of several named individuals are telling the truth and
which are lying \citep{knights_knaves}. It is a challenging task that tests both logical reasoning and generalization to novel statement structures; we use the \textsc{Reasoning Gym} \citep{reasoning_gym} implementation.

An instance of the task describes an island inhabited only by two named individuals. Each of them is either a knight, who always tells the truth, or a knave, who always lies. Different instances use different equivalent terminologies: e.g., knight/knave or hero/villain. Each individual makes one statement about the types of the individuals on the island; statements can be simple assertions or nested combinations of negation, conjunction, disjunction, implication, and biconditional over other individuals' statements. Every instance is guaranteed to have a unique assignment of types. Task variability comes from the sampled statements, the names used, and the terminology.

The model needs to respond with a type for each of the two individuals. A task is considered solved iff the assigned types match the unique solution consistent with the statements.

\subsection{\textsc{Sudoku}}
This is the $4\times4$ variant of Sudoku, which is a constraint-satisfaction task on which language models continue to struggle even when the board size is reduced \citep{sudoku_bench}. We use the \textsc{Reasoning Gym} \citep{reasoning_gym} implementation at the hardest difficulty setting.

An instance of the task is a $4\times4$ grid, which is partially filled with digits between $1$ to $4$, such that there is a unique completion that satisfies the following constraints: each of the four rows, each of the four columns, and each of the four non-overlapping $2\times2$ subgrids must contain every digit from $1$ to $4$ exactly once. At the difficulty setting that we use, $12$ of the $16$ cells are empty initially, which is the maximum number of empty cells for which a unique solution can be guaranteed on a $4\times4$ board. This means that that only $4$ clues are given. Task variability comes from the underlying solved grid and from which $12$ cells are removed to form the starting state.

The model needs to respond with a completed $4\times4$ grid. A task is considered solved iff the completed grid exactly matches the (unique) solution.

\subsection{\textsc{GSM-Symbolic}}
This is a procedurally generated variant of GSM8K \citep{gsm8k} that is designed to test mathematical reasoning more reliably than the original benchmark \citep{gsm_symbolic}: models frequently perform worse on \textsc{GSM-Symbolic} than on the corresponding original GSM8K problems, which suggests that some of their performance on GSM8K is due to memorization. We use the \textsc{Reasoning Gym} \citep{reasoning_gym} implementation.

An instance of the task is generated by taking one of a fixed set of GSM8K-style word-problem templates and resampling its numeric quantities and entity names. Task variability comes from both the choice of template and the resampled quantities and names.

The model needs to respond with a single (final) numeric answer to the word problem. A task is considered solved iff this value exactly matches the oracle answer for the sampled instance.
    % !TEX root =  main.tex
%%%%%%%%%%%%%%%%%%%%%%%%%%%%%%%%%%%%%
\section{Training Details}\label{sec:app.training_details}

\subsection{\textsc{Arithmetic}}
For this task, we use a small causal transformer: \texttt{TinyGPT} with 2 layers and 4 attention heads (more details in \citep{upskill}). \textsc{Single} uses a single such network, and \textsc{Stratified} instantiates $k=5$ separate TinyGPT networks. We train both methods from a random initialization, with supervised teacher-forcing warmup phase for 100 steps and using RL for pass@5 for 2000 steps.

\subsection{\textsc{Knights-\&-Knaves}}
Both \textsc{Single} and \textsc{Stratified} fine-tune Qwen2.5-1.5B-Instruct \citep{qwen} with a learning rate of $5\times10^{-6}$. We first train a single LoRA adapter for 2000 steps, and then use it to initialize the adapters for these methods, which are trained for a further 2000 steps.

\subsection{\textsc{Sudoku}}
Both \textsc{Single} and \textsc{Stratified} fine-tune Qwen2.5-1.5B-Instruct with a learning rate of $1\times10^{-6}$. We first train a single LoRA adapter on an easy tier of $4\times4$ \textsc{Sudoku} (4 to 8 empty cells) for 20{,}000 steps, then continue training it on a harder tier (8 to 10 empty cells) for a further 20{,}000 steps. We use this adapter to initialize \textsc{Single} and \textsc{Stratified}, which are trained for 10{,}000 steps at the hardest difficulty level (12 out of 16 cells empty).

\textbf{Independently trained ensemble baseline.} \textsc{Independent} (Fig.~\ref{fig:additional_results}~(c)) is a population of five independently trained \textsc{Single} policies. Each is initialized from the same warm-start adapter and uses the same configuration as \textsc{Single}, but is trained for only $2{,}000$ steps to ensure that the five runs match the compute of one \textsc{Stratified} run ($5 \times 2{,}000 = 10{,}000$ steps). We train $15$ policies in three populations of five and report the mean and standard error over the three populations. For pass@$5$, we take one sample from each policy in a population, similar to \textsc{Stratified}.

\subsection{\textsc{GSM-Symbolic}}
Both \textsc{Single} and \textsc{Stratified} fine-tune Llama-3.1-8B-Instruct \citep{llama} with a learning rate of $1\times10^{-6}$. They are trained without any warm-start for 2000 steps.

% !TEX root =  main.tex
\begin{table}[h]
\centering
\caption{GRPO training hyperparameters.}
\label{tab:app.shared_hparams}
\begin{tabular}{ll}
\toprule
\textbf{Hyperparameter} & \textbf{Value} \\
\midrule
LoRA Rank                        & 32 (160 for \textsc{Big-Single}) \\
Population Size $k$ (\texttt{num\_generations}) & 5 \\
Training Batch Size              & 5 (one prompt per training step) \\
Training Epochs                  & 1 \\
Optimizer                        & AdamW (8-bit) \\
Adam $\beta_1$, $\beta_2$        & 0.9, 0.99 \\
Weight Decay                     & 0.0 \\
Gradient Accumulation Steps      & 1 \\
Max Gradient Norm                & 1.0 \\
Learning-rate Schedule           & constant \\
KL Coefficient ($\beta$)         & 0.0 \\
Sampling Temperature             & 1.0 \\
Max Prompt Length                & 512 tokens \\
Max Completion Length            & 768 tokens \\
Max Sequence Length              & 1024 tokens \\
Precision                        & bf16 \\
\bottomrule
\end{tabular}
\end{table}
    % !TEX root =  main.tex
%%%%%%%%%%%%%%%%%%%%%%%%%%%%%%%%%%%%%
\section{Evaluation Details}\label{sec:app.eval_details}

\textbf{Evaluation Dataset.}
For \textsc{Knights-\&-Knaves}, \textsc{Sudoku}, and \textsc{GSM-Symbolic}, we evaluate on the same $1000$ randomly sampled instances for all methods and seeds, and on $300$ randomly sampled problems for \textsc{Arithmetic}.

\textbf{Decoding.}
For the LLM tasks, responses are sampled with vLLM at the same sampling parameters as the training rollouts: temperature $1.0$, top-$p=1.0$, no top-$k$ truncation, and a maximum of $768$ new tokens. For \textsc{Arithmetic}, responses are sampled at temperature $0.9$ (see \citep{upskill}).

\textbf{Checkpoint.}
We evaluate the last checkpoint of each run for \textsc{Arithmetic}, \textsc{Knights-\&-Knaves}, and \textsc{Sudoku}. For \textsc{GSM-Symbolic}, \textsc{Single} collapses late in training (see Fig.~\ref{fig:additional_results}b), so we report the last checkpoint before the collapse based on training rewards.

\textbf{Diversity Analyses.}
The overlap analyses (Figs.~\ref{fig:jaccard_heatmaps}~and~\ref{fig:unique_solves}) use greedy decoding, on $1000$ instances for \textsc{Sudoku} and \textsc{GSM-Symbolic} and $500$ for \textsc{Knights-\&-Knaves} and \textsc{Arithmetic}.
    % !TEX root =  main.tex
%%%%%%%%%%%%%%%%%%%%%%%%%%%%%%%%%%%%%
\section{Full Results}\label{sec:app.full_results}

% !TEX root =  main.tex
\begin{table*}[h]
    \caption{Full results in \% with mean (standard error) over three seeds.}
    \label{tab:app.full_results}
    \centering
    \scalebox{0.78}{
        \setlength\tabcolsep{5pt}
        \renewcommand{\arraystretch}{1.12}
        \begin{tabular}{lcccccccc}
            \toprule
            & \multicolumn{2}{c}{\textsc{Arithmetic}} & \multicolumn{2}{c}{\textsc{Knights-\&-Knaves}} & \multicolumn{2}{c}{\textsc{Sudoku}} & \multicolumn{2}{c}{\textsc{GSM-Symbolic}} \\
            \cmidrule(lr){2-3}\cmidrule(lr){4-5}\cmidrule(lr){6-7}\cmidrule(lr){8-9}
            \textbf{Method} & Pass@1 & Pass@5 & Pass@1 & Pass@5 & Pass@1 & Pass@5 & Pass@1 & Pass@5 \\
            \midrule
            \textsc{Single} & $93.4$ ($3.9$) & $95.5$ ($2.8$) & $56.8$ ($0.8$) & $90.6$ ($0.6$) & $40.9$ ($0.3$) & $74.2$ ($0.6$) & $52.1$ ($4.0$) & $85.5$ ($2.4$) \\
            \midrule
            \textsc{Stratified}, adapter 1 & $62.9$ ($3.7$) & -- & $36.6$ ($6.4$) & -- & $34.5$ ($0.8$) & -- & $56.2$ ($0.5$) & -- \\
            \textsc{Stratified}, adapter 2 & $40.2$ ($0.4$) & -- & $32.5$ ($3.2$) & -- & $34.8$ ($1.7$) & -- & $54.9$ ($1.1$) & -- \\
            \textsc{Stratified}, adapter 3 & $23.5$ ($4.1$) & -- & $41.4$ ($0.9$) & -- & $35.5$ ($0.5$) & -- & $56.9$ ($0.7$) & -- \\
            \textsc{Stratified}, adapter 4 & $20.1$ ($1.1$) & -- & $45.0$ ($0.2$) & -- & $34.0$ ($0.7$) & -- & $56.9$ ($0.6$) & -- \\
            \textsc{Stratified}, adapter 5 & $15.8$ ($10.4$) & -- & $43.2$ ($2.8$) & -- & $34.4$ ($1.1$) & -- & $58.1$ ($1.2$) & -- \\
            \midrule
            \textsc{Stratified} & $32.5$ ($1.7$) & $99.7$ ($0.4$) & $39.8$ ($0.7$) & $93.2$ ($0.8$) & $34.6$ ($0.5$) & $80.3$ ($0.3$) & $56.6$ ($0.2$) & $90.4$ ($0.1$) \\
            \bottomrule
        \end{tabular}
    }
\end{table*}

We provide the full results for our main experiments in Table~\ref{tab:app.full_results}. Pass@$1$ is defined as the expected accuracy of a single response; for \textsc{Single}, it is the mean accuracy over its samples, and for \textsc{Stratified}, it is the mean accuracy over its five adapters (which is equivalent to the accuracy of one response sampled from an adapter chosen uniformly at random).
    % !TEX root =  main.tex
%%%%%%%%%%%%%%%%%%%%%%%%%%%%%%%%%%%%%
\rev{
\section{Choice of Advantage Baseline}\label{sec:app.baseline}

% !TEX root =  main.tex
\begin{table*}[h]
    \caption{\rev{Comparison between group-mean (Eq.~\ref{eq:mean_wta_advantage}) and leave-one-out (Eq.~\ref{eq:loo_advantage}) advantage.}}
    \label{tab:app.loo_ablation}
    \centering
    \scalebox{0.85}{
        \setlength\tabcolsep{5pt}
        \renewcommand{\arraystretch}{1.12}
        \rev{
        \begin{tabular}{llcccccccc}
            \toprule
            & & \multicolumn{4}{c}{\textbf{Group-Mean}} & \multicolumn{4}{c}{\textbf{Leave-One-Out}} \\
            \cmidrule(lr){3-6}\cmidrule(lr){7-10}
            & \textbf{Method} & Seed 1 & Seed 2 & Seed 3 & Mean & Seed 1 & Seed 2 & Seed 3 & Mean \\
            \midrule
            \multirow{3}{*}{\rotatebox[origin=c]{90}{\small\textsc{Sudoku}}}
            & \textsc{Single}     & $75.0$ & $73.1$ & $74.4$ & $74.2$ ($0.6$) & $72.5$ & $75.3$ & $72.4$ & $73.4$ ($1.0$) \\
            & \textsc{Stratified} & $80.0$ & $80.0$ & $80.8$ & $80.3$ ($0.3$) & $74.2$ & $76.1$ & $77.5$ & $75.9$ ($1.0$) \\
            & Gap                 & $+5.0$ & $+6.9$ & $+6.4$ & $+6.1$ ($0.6$) & $+1.7$ & $+0.8$ & $+5.1$ & $+2.5$ ($1.3$) \\
            \midrule
            \multirow{3}{*}{\rotatebox[origin=c]{90}{\small\textsc{K\&K}}}
            & \textsc{Single}     & $91.2$ & $91.2$ & $89.4$ & $90.6$ ($0.6$) & $88.9$ & $84.8$ & $85.9$ & $86.5$ ($1.2$) \\
            & \textsc{Stratified} & $94.7$ & $92.7$ & $92.1$ & $93.2$ ($0.8$) & $91.9$ & $85.0$ & $88.5$ & $88.5$ ($2.0$) \\
            & Gap                 & $+3.5$ & $+1.5$ & $+2.7$ & $+2.6$ ($0.6$) & $+3.0$ & $+0.2$ & $+2.6$ & $+1.9$ ($0.9$) \\
            \bottomrule
        \end{tabular}
        }
    }
\end{table*}

Table~\ref{tab:app.loo_ablation} compares the group-mean baseline with the leave-one-out baseline on \textsc{Sudoku} and \textsc{Knights-\&-Knaves}. The results show two things. First, the group-mean baseline improves \textsc{Stratified} on every seed of both tasks ($+4.3$ on average on \textsc{Sudoku}, $+4.7$ on \textsc{Knights-\&-Knaves}). Its effect on \textsc{Single} is small and not consistent across seeds for \textsc{Sudoku}, but it improves \textsc{Single} on every seed for \textsc{Knights-\&-Knaves}. Second, \textsc{Stratified} outperforms \textsc{Single} on all seeds under both baselines, but the gap is smaller under leave-one-out.

Additionally, Fig.~\ref{fig:update_counts} shows that the group-mean baseline provides a denser learning signal than LOO, with all
adapters receiving gradient updates throughout training.

% !TEX root =  main.tex
\begin{figure*}[h]
    \centering
    \includegraphics[
        width=\textwidth
    ]{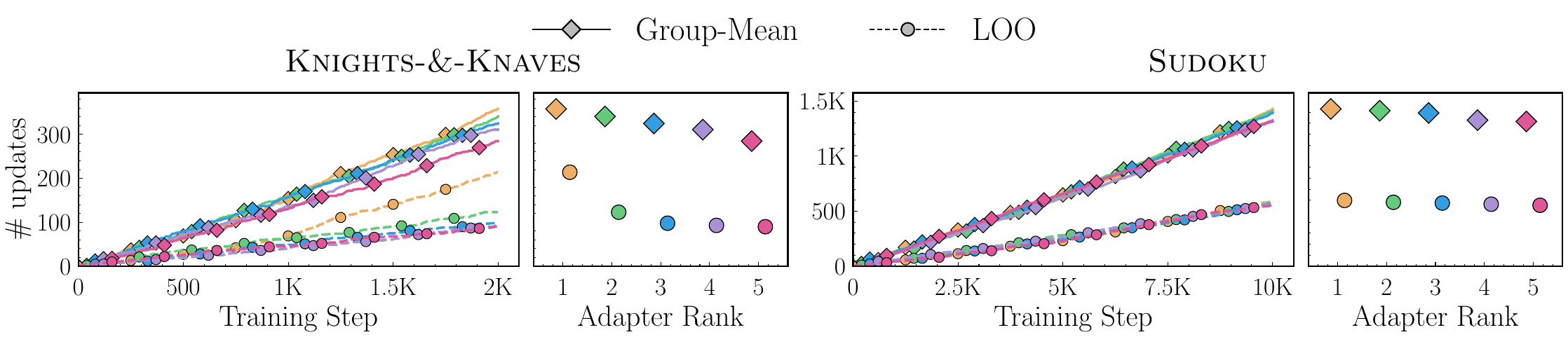}
    \vspace{-2em}
    \caption{\rev{Cumulative number of updates per adapter during training for the group-mean (Eq.~\ref{eq:mean_wta_advantage}) and LOO (Eq.~\ref{eq:loo_advantage}) advantages. For each task, the panel on the right shows the final update counts for adapters ranked by final counts. Group-mean provides a denser learning signal, which leads to more updates for every adapter, and the updates are also spread relatively evenly across adapters.}}
    \label{fig:update_counts}
\end{figure*}

Overall, these results show that stratification improves performance under both the pass@$k$ objective optimized by LOO and the objective optimized by group-mean, which we analyze in Appendix~\ref{sec:app.maxrl}.
}
    % !TEX root =  main.tex
%%%%%%%%%%%%%%%%%%%%%%%%%%%%%%%%%%%%%
%%%%%%%%%%%%%%%%%%%%%%%%%%%%%%%%%%%%%
\rev{
\section{Complementarity Analysis}
\label{sec:app.complementarity}

% !TEX root =  main.tex
\begin{figure*}[h]
    \centering
    \includegraphics[
        width=\textwidth
    ]{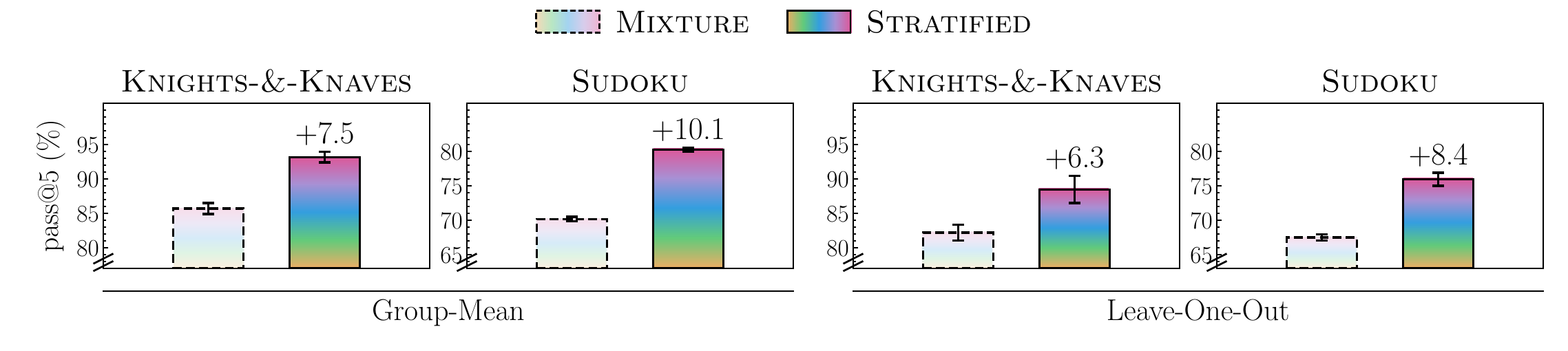}
    \vspace{-2em}
    \caption{\rev{Pass@$5$ accuracy of \textsc{Mixture} versus \textsc{Stratified} under the group-mean (Eq.~\ref{eq:mean_wta_advantage}) and LOO (Eq.~\ref{eq:loo_advantage}) advantages on \textsc{Knights-\&-Knaves} and \textsc{Sudoku} tasks. Error bars denote standard error across three seeds. Each panel's $y$-axis is truncated (kink mark) to make the gap visible. There is a substantial gap between \textsc{Stratified} and \textsc{Mixture} on both tasks under both estimators.}}
    \label{fig:mixture_gap}
\end{figure*}

Let $p_i(x)$ denote the probability that a response sampled from policy $\pi_{\theta_i}(\cdot\mid x)$ is correct. \textsc{Stratified} independently samples one response from each of the $k$ policies. In contrast, \textsc{Mixture} samples $k$ responses independently from the uniform mixture $\bar\pi(\cdot\mid x)=\frac{1}{k}\sum_{i=1}^{k}\pi_{\theta_i}(\cdot\mid x)$.
Define $\bar p(x) \coloneqq \frac{1}{k}\sum_{i=1}^{k}p_i(x)$. The probabilities that at least one of $k$ responses is correct are
\[
P_{\textsc{Stratified}}(x)
=1-\prod_{i=1}^{k}(1-p_i(x)),
\qquad
P_{\textsc{Mixture}}(x)
=1-(1-\bar p(x))^k.
\]
By the AM-GM inequality, the following holds:
\[
\prod_{i=1}^{k}(1-p_i(x))
\leq
\left(\frac{1}{k}\sum_{i=1}^{k}(1-p_i(x))\right)^k
=(1-\bar p(x))^k.
\]
The gap $P_{\textsc{Stratified}}(x)-P_{\textsc{Mixture}}(x)$ is zero iff all the policies have the same success probability on $x$. Fig.~\ref{fig:mixture_gap} shows its expectation over the prompt distribution as a measure of complementarity.

To further analyze if a single policy in the population is responsible for \textsc{Stratified}'s performance, we evaluate the pass@$5$ accuracy of each adapter individually, using five independent samples from that adapter. Table~\ref{tab:app.adapter_pass5} shows that even the best adapter (selected in hindsight) underperforms both \textsc{Single} and \textsc{Stratified} on every seed of both tasks.

% !TEX root = main.tex
\begin{table*}[h]
    \caption{\rev{Pass@$5$ accuracy of individual LoRA adapters in our stratified population of five specialized policies,
    compared to \textsc{Single} and \textsc{Stratified}, with adapters ranked by their pass@$5$ accuracy. Even the best LoRA adapter in the population underperforms both \textsc{Single} and \textsc{Stratified}.}}
    \label{tab:app.adapter_pass5}
    \centering
    \scalebox{0.85}{
        \setlength\tabcolsep{5pt}
        \renewcommand{\arraystretch}{1.12}
        \rev{
        \begin{tabular}{llccccccc}
            \toprule
            & & \multicolumn{5}{c}{\textbf{Individual Adapters in Stratified Population}}
              & \multicolumn{2}{c}{\textbf{Method}} \\
            \cmidrule(lr){3-7}\cmidrule(lr){8-9}
            & Seed & Rank 1 & Rank 2 & Rank 3 & Rank 4 & Rank 5
            & \textsc{Single} & \textsc{Stratified} \\
            \midrule
            \multirow{3}{*}{\rotatebox[origin=c]{90}{\small\textsc{Sudoku}}}
            & 1 & $46.2$ & $42.6$ & $41.8$ & $39.6$ & $37.7$ & $75.0$ & $80.0$ \\
            & 2 & $45.7$ & $42.0$ & $40.9$ & $40.6$ & $40.6$ & $73.1$ & $80.0$ \\
            & 3 & $48.1$ & $44.6$ & $40.8$ & $40.3$ & $38.0$ & $74.4$ & $80.8$ \\
            \midrule
            \multirow{3}{*}{\rotatebox[origin=c]{90}{\small\textsc{K\&K}}}
            & 1 & $69.1$ & $57.1$ & $54.8$ & $51.5$ & $39.4$ & $91.2$ & $94.7$ \\
            & 2 & $67.0$ & $65.2$ & $61.9$ & $60.9$ & $35.0$ & $91.2$ & $92.7$ \\
            & 3 & $69.2$ & $64.9$ & $62.1$ & $61.7$ & $37.4$ & $89.4$ & $92.1$ \\
            \bottomrule
        \end{tabular}
        }
    }
\end{table*}

\par
% !TEX root = main.tex
\begin{wraptable}{r}{0.40\textwidth}
    \centering
    \vspace{-1.2em}
    \caption{\rev{Coverage of the best subset of learned LoRA adapters on
    \textsc{Knights-\&-Knaves} using greedy decoding. The gap is w.r.t. the full population.}}
    \label{tab:app.subset_coverage}
    \scalebox{0.85}{
        \setlength{\tabcolsep}{5pt}
        \renewcommand{\arraystretch}{1.12}
        \rev{
        \begin{tabular}{ccc}
            \toprule
            Subset Size & Best Coverage & Gap \\
            \midrule
            2 & $74.6\%$ & $19.0$ \\
            3 & $84.6\%$ & $9.0$ \\
            4 & $90.8\%$ & $2.8$ \\
            5 & $93.6\%$ & $0.0$ \\
            \bottomrule
        \end{tabular}
        }
    }
\end{wraptable}
Although the bimodal pattern in the Jaccard similarity heatmap (see Fig.~\ref{fig:jaccard_heatmaps}) for
\textsc{Knights-\&-Knaves} shows that some policies solve similar sets
of problems, it does not necessarily imply that two policies are
sufficient. To make this more concrete, we evaluate the sets of solved
problems for every subset of the five policies on
\textsc{Knights-\&-Knaves} using greedy decoding.
Table~\ref{tab:app.subset_coverage} reports the best subset's coverage
for each subset size, along with the gap with respect to using the
full population. Even the best pair of adapters (selected in
hindsight) is severely suboptimal compared to the full population.
\par
}
    % !TEX root =  main.tex
%%%%%%%%%%%%%%%%%%%%%%%%%%%%%%%%%%%%%
%%%%%%%%%%%%%%%%%%%%%%%%%%%%%%%%%%%%%
\rev{
\vspace{-0.8em}
\section{Effect of Population Size}
\label{sec:app.population_size}

We evaluate \textsc{Single} and \textsc{Stratified} on \textsc{Sudoku} with varying population size $k\in\{3,5,8\}$, and train using the group-mean advantage. Fig.~\ref{fig:population_size} shows that \textsc{Stratified} beats \textsc{Single} at all three population sizes ($+6.2$, $+6.1$, and $+2.2$ improvement in pass@$k$, respectively). Further, the improvement is positive on every seed at each value of $k$.

% !TEX root =  main.tex
\begin{figure}[h]
    \centering
    \vspace{-1em}
    \includegraphics[width=0.5\textwidth]{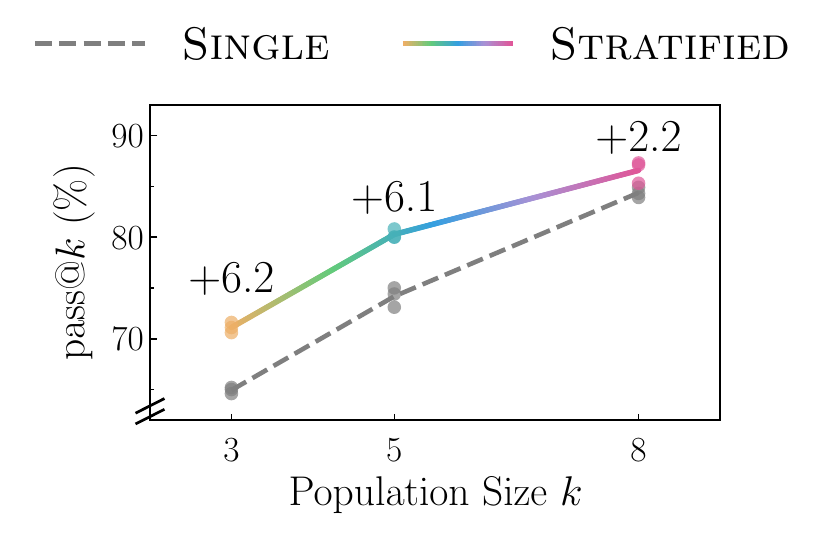}
    \vspace{-1em}
    \caption{\rev{Pass@$k$ accuracy (averaged across three seeds) on \textsc{Sudoku} for different
    population sizes $k$, with results for individual seeds shown as faint dots. \textsc{Stratified} consistently beats \textsc{Single}.}}
    \label{fig:population_size}
    \vspace{-1em}
\end{figure}
}
    % !TEX root =  main.tex
%%%%%%%%%%%%%%%%%%%%%%%%%%%%%%%%%%%%%
%%%%%%%%%%%%%%%%%%%%%%%%%%%%%%%%%%%%%
\rev{
\section{Connection to Maximum Likelihood RL}\label{sec:app.maxrl}
We first characterize the learning objective that the group-mean advantage optimizes for (Section~\ref{sec:gm_obj}). Section~\ref{sec:maxrl} then connects it to Maximum Likelihood RL (\textsc{MaxRL}) \citep{maxrl}.

\subsection{Objective of the Group-Mean Baseline}
\label{sec:gm_obj}

\textbf{Notation.} Let $x \sim \rho$ denote a prompt sampled from the prompt distribution $\rho$. The population comprises $k$ policies $\pi_{\theta_i}$ with parameters $\Theta = (\theta_1,\ldots,\theta_k)$. Given $x$, we independently sample $N=k$ responses, one response $y_i \sim \pi_{\theta_i}(\cdot \mid x)$ from each policy, and assign them binary rewards $r_i = r(x,y_i) \in \{0,1\}$. The winner $i^\star$ is sampled uniformly from $\arg\max_i r_i$, and each response is assigned advantage $A_i = \mathbf{1}\{i=i^\star\} \left(r_i-\frac{1}{k}\sum_{j=1}^{k}r_j\right)$. Similar to the convention in \citet{maxrl}, $N$ denotes the number of rollouts, $K = \sum_{i=1}^{k}r_i$ is the number of successes, $T \in \mathbb{N}$ is the truncation level for the log-likelihood expansion, and $S_i \coloneqq \nabla_{\theta_i} \log m_{\theta_i}(z_i \mid x)$ denotes the score function. Response $y_i$ and policy $\pi_\theta$ in our setting correspond to $z_i$ and $m_\theta$ in \citet{maxrl}. 

\begin{prop}[Objective of group-mean advantage]
Let $K = \sum_{i=1}^{k} r_i$ denote the number of successes and
$\bar r = K/k$ the empirical success rate. Define the utility
\[
F_{\textsc{GM}}(x,y_1,\ldots,y_k)
\coloneqq \sum_{m=1}^{K}\left(\frac{1}{m}-\frac{1}{k}\right),
\]
where the sum is defined to be zero when $K=0$, and the objective
\[
J_{\textsc{GM}}(\Theta)
\coloneqq \mathbb{E}_{x\sim\rho,\;
y_{1:k}\sim\bigotimes_{i=1}^{k}\pi_{\theta_i}(\cdot\mid x)}
\left[F_{\textsc{GM}}(x,y_1,\ldots,y_k)\right].
\]
Let $A_i = \mathbf{1}\{i=i^\star\}(r_i-\bar r)$ be the group-mean advantage in Eq.~\ref{eq:mean_wta_advantage}. Then the following holds for every policy $i\in\{1,\ldots,k\}$:
\[
\mathbb{E}\left[A_i S_i\right]
= \nabla_{\theta_i}J_{\textsc{GM}}(\Theta),
\]
where the expectation includes the randomness in selecting $i^\star$.
\label{prop:group-mean-objective}
\end{prop}

\textit{Proof.}
Consider the leave-one-out (LOO) baseline
\[
b_i \coloneqq \sum_{m=1}^{K-r_i}\left(\frac{1}{m}-\frac{1}{k}\right).
\]
The advantage equals $\widetilde A_i \coloneqq F_{\textsc{GM}}-b_i = r_i(1/K-1/k)$ for $K>0$, and zero otherwise.

When $K=0$, both $A_i$ and $\widetilde A_i$ are zero. When $K>0$, response $i$ is selected as the winner with probability $r_i/K$. Therefore,
\[
\mathbb{E}_{i^\star}[A_i\mid x,y_{1:k}]
= \frac{r_i}{K}\left(1-\frac{K}{k}\right)
= r_i\left(\frac{1}{K}-\frac{1}{k}\right)
= \widetilde A_i.
\]

Consequently, the following holds:
\[
\begin{aligned}
\mathbb{E}[A_i S_i]
&= \mathbb{E}\!\left[
    \mathbb{E}_{i^\star}[A_i\mid x,y_{1:k}]S_i
    \right]\\
&= \mathbb{E}[(F_{\textsc{GM}}-b_i)S_i]\\
&= \mathbb{E}[F_{\textsc{GM}}S_i]\\
&= \nabla_{\theta_i}J_{\textsc{GM}}(\Theta).
\end{aligned}
\]

This completes the proof.

\textbf{Remark.} The $m$-th success increments $F_{\textsc{GM}}$ by \(1/m-1/k\), i.e., additional successes are rewarded with decreasing returns. In contrast, the best-of-\(k\) utility rewards only the first success.

\subsection{Relation to the MaxRL Objective}
\label{sec:maxrl}

\begin{prop}[Connection to \textsc{MaxRL}]
Consider the shared parameter setting $\theta_1=\cdots=\theta_k=\theta$, where all $k$ responses are sampled independently from $\pi_\theta$. Define the pass rate and RL objective
\[
p_\theta^{\mathrm{pass}}(x)
=
\mathbb E_{y\sim\pi_\theta(\cdot\mid x)}[r(x,y)],
\qquad
J_{\textsc{RL}}(\theta)
=
\mathbb E_{x\sim\rho}[p_\theta^{\mathrm{pass}}(x)].
\]
Similar to \textsc{MaxRL} \citep{maxrl}, define the truncated log-likelihood objective
\[
J_{\textsc{MaxRL}}^{(T)}(\theta)
=
-\mathbb E_{x\sim\rho}
\left[
\sum_{t=1}^{T}
\frac{(1-p_\theta^{\mathrm{pass}}(x))^t}{t}
\right].
\]

Then the following holds:
\[
\nabla_\theta J_{\textsc{GM}}
=
\nabla_\theta J_{\textsc{MaxRL}}^{(k)}
-
\nabla_\theta J_{\textsc{RL}}.
\]
\end{prop}

\textit{Proof.}
Fix a prompt \(x\). The unclipped REINFORCE-style \textsc{MaxRL} estimator is
\[
\widehat g_N
=
\begin{cases}
\displaystyle\frac1K\sum_{i=1}^{N}r_iS_i,&K>0,\\[5pt]
0,&K=0.
\end{cases}
\]

Our group-mean estimator is \(\widehat g_{\textsc{GM}}=\sum_i A_iS_i\). Therefore,
\[
\begin{aligned}
\mathbb E_{i^\star}[\widehat g_{\textsc{GM}}\mid x,y_{1:k}]
&=\left(1-\frac Kk\right)\widehat g_k\\
&=\widehat g_k-\frac1k\sum_{i=1}^k r_iS_i,
\end{aligned}
\]
where the second term is the standard REINFORCE estimator. Taking expectations over the responses and prompts, and using the unbiasedness of $\widehat g_k$ for $\nabla_\theta J_{\textsc{MaxRL}}^{(k)}$ \citep{maxrl}, along with Proposition~\ref{prop:group-mean-objective}, completes the proof.

\textbf{Remark.}  \citet{maxrl} show that the truncated \textsc{MaxRL} objective has the same gradient as a weighted harmonic sum of pass@$j$ objectives:
\[
\nabla_\theta J_{\textsc{MaxRL}}^{(k)}(\theta)
=
\nabla_\theta \mathbb{E}_{x\sim\rho}\left[
\sum_{j=1}^{k}\frac{1}{j}\operatorname{pass}@j(x)
\right].
\]
$J_{\textsc{RL}}$ is the expected pass@$1$ objective, which gives us the following:
\[
\nabla_\theta J_{\textsc{GM}}(\theta)
=
\nabla_\theta \mathbb{E}_{x\sim\rho}\left[
\sum_{j=2}^{k}\frac{1}{j}\operatorname{pass}@j(x)
\right].
\]
This shows that the updates using the group-mean baseline remove the pass@$1$ term in this setting. \citet{maxrl} empirically show that \textsc{MaxRL} beats pass@$k$ optimization, which is consistent with our experiments, where the group-mean baseline performs better than LOO.
}
}

\end{document}